\pdfoutput=1

\documentclass[11pt]{article}

\usepackage{ACL2023}

\usepackage{times}
\usepackage{latexsym}

\usepackage[T1]{fontenc}

\usepackage[utf8]{inputenc}

\usepackage{microtype}

\usepackage{inconsolata}

\usepackage{lipsum}  
\usepackage{tabularx}
\usepackage{booktabs}

\usepackage{microtype}
\usepackage{type1cm}
\usepackage[american]{babel}
\usepackage{amssymb}
\usepackage{multirow}
\usepackage{xspace}
\DeclareMathAlphabet{\mathcalbf}{OMS}{pzc}{b}{n}
\usepackage{enumitem}
\usepackage{colortbl}

\RequirePackage{color}
\definecolor{darkgray}{gray}{0.40}
\definecolor{mediumgray}{gray}{0.60}
\definecolor{lightgray}{gray}{0.95}
\definecolor{ultralightgray}{gray}{0.98}
\definecolor{forestgreen}{rgb}{0.133, 0.545, 0.133}
\definecolor{orange}{rgb}{1, 0.86, 0.74}
\definecolor{lightergreen}{rgb}{0.95, 1, 0.88}

\usepackage{graphicx}
\DeclareGraphicsExtensions{.pdf,.ai,.jpg,.png}
\setkeys{Gin}{pagebox=artbox}
\graphicspath{{../acl23-appropriateness-corpus-figures/}}

\newcommand{\bsfigure}[3][]{%
    \begin{figure}[t]
        \centering
        \includegraphics[#1]{#2}
        \caption{#3}\label{#2}%
    \end{figure}
}

\newcommand{\hwfigure}[3][t!]{%
    \begin{figure*}[#1]
        \centering
        \includegraphics[scale=1.0]{#2}
        \caption{#3}\label{#2}%
    \end{figure*}
}

\RequirePackage{type1cm}
\RequirePackage{color}
\RequirePackage{soul}
\setstcolor{blue}
\definecolor{violet}{rgb}{0.5,0.0,0.5}

\newsavebox\bscombox
\newcommand{\bscom}[3][]{%
    \sbox{\bscombox}{\fontsize{8}{9}\selectfont#1#2#3}
    \noindent
    \st{#2}{\selectfont
        \color{blue}#3\ifx\\#1\\\else{\fontsize{8}{9}\selectfont\color{violet}[#1]}\fi
    }
}

\begin{document}

\title{Modeling Appropriate Language in Argumentation}

\setlength\titlebox{6.5cm}
\author{
  Timon Ziegenbein \\
  Leibniz University Hannover \\
  \texttt{t.ziegenbein@ai.uni-hannover.de} \And
  Shahbaz Syed\\
  Leipzig University \\
  \texttt{shahbaz.syed@uni-leipzig.de} \\\AND
  Felix Lange\\
  Paderborn University\\
  \texttt{flange@mail.uni-paderborn.de} \\\And
  Martin Potthast\\
  Leipzig University \\
  \texttt{martin.potthast@uni-leipzig.de} \\\AND
  Henning Wachsmuth\\
  Leibniz University Hannover \\
  \texttt{h.wachsmuth@ai.uni-hannover.de} \\\
}

\maketitle

\begin{abstract}
Online discussion moderators must make ad-hoc decisions about whether the contributions of discussion participants are \emph{appropriate} or should be removed to maintain civility. Existing research on offensive language and the resulting tools cover only one aspect among many involved in such decisions. The question of what is considered appropriate in a controversial discussion has not yet been systematically addressed. In this paper, we operationalize appropriate language in argumentation for the first time. In particular, we model appropriateness through the absence of flaws, grounded in research on argument quality assessment, especially in aspects from rhetoric. From these, we derive a new taxonomy of 14~dimensions that determine inappropriate language in online discussions. Building on three  argument quality corpora, we then create a corpus of 2191~arguments annotated for the 14~dimensions. Empirical analyses support that the taxonomy covers the concept of appropriateness comprehensively, showing several plausible correlations with argument quality dimensions. Moreover, results of baseline approaches to assessing appropriateness suggest that all dimensions can be modeled computationally on the corpus.
\end{abstract}

\section{Introduction}
\label{sec:introduction}

\bsfigure{appropriateness-example}{Two arguments from the corpus introduced in this paper, one appropriate and one inappropriate. The used colors match the taxonomy concepts we present in Section~\ref{sec:modeling}: toxic intensity (dark red), unclear meaning (orange), and missing openness (light purple).}

People have varying degrees of sensitivity to controversial issues and may be triggered by different emotional responses dependent on the issue and the opponents' arguments \cite{walton:2010}. This often makes it hard to maintain a constructive discussion. In competitive debates, a moderator ensures that participants argue \emph{appropriately}. Debating culture, dating back to the 18th century, demands appropriate behavior, such as staying on topic and avoiding overly emotional language \cite{andrew:1996}. Accordingly, \citet{wachsmuth:2017a} define arguments to be appropriate if they support credibility and emotions and match the issue.

Similarly, in many online forums, moderators ensure a certain level of civility in the discussions. What arguments are considered civil may differ from community to community. The task of discussion moderation thus requires ad-hoc decisions about the appropriateness of any contributed argument, calling out the inappropriate ones---a challenging task to master. Moreover, the amount of moderation required on the web necessitates automation of this task, as the resources for manual moderation are usually insufficient.

Figure~\ref{appropriateness-example} shows two exemplary arguments, assessed by human annotators. The inappropriate argument appeals excessively to emotions, is not easily understandable, and shows little interest in the opinion of others. Note that the last sentence of the argument is also a personal attack, a special case of inappropriate emotional language. Hence, multiple inappropriateness aspects can occur at the same time. The appropriate argument, on the other hand, does not contain any of these issues. 

Most previous work on automatic content moderation has focused on detecting offensive content \cite{schmidt:2017,poletto:2021}. However, to create a climate in which controversial issues can be discussed constructively, combating only offensive content is not enough, since there are also many other forms of inappropriate arguments \cite{habernal:2018}. While the notion of appropriateness is treated in argumentation theory as an important subdimension of argument quality (see Section~\ref{sec:related-work}), there has been no systematic study of appropriateness, let alone a clear definition or operationalization. These shortcomings hinder the development of automatic moderation tools.

In this paper, we present a taxonomy of 14~inappropriateness dimensions, systematically derived from rhetoric \cite{burkett:2011} and argument quality theory \cite{wachsmuth:2017a}, along with a corpus annotated for the dimensions. Matching elements of the concept of reasonableness by \citet{vaneemeren:2015}, we argue appropriateness to be a minimal quality property that is necessary for any argument to consider it valuable in a debate. 

We motivate the 14~dimensions empirically in Section~\ref{sec:modeling} by analyzing interactions of low appropriateness with other quality issues of arguments, and we further refine the dimensions on this basis. To operationalize the taxonomy, we create a new corpus of 2191~arguments from debates, question-answering forums, and reviews (Section \ref{sec:corpus}). The arguments are compiled from three existing argument quality corpora \cite{habernal:2016b,wachsmuth:2017a,ng:2020}, such that they cover both a variety of topics and selected topics in depth. All arguments are manually labeled for the dimensions in a human annotation study. 

Given the new corpus, we analyze correlations between the 14~dimensions and the argument quality dimensions in the source corpora in Section~\ref{sec:analysis}. Several plausible correlations support that our taxonomy successfully aligns with the theoretical and practical quality aspects modeled in previous work. To gain insights into how well the proposed dimensions can be predicted automatically, we also evaluate first baseline approaches to the computational assessment of appropriateness (Section~\ref{sec:experiments}). The results do not fully compete with the average human performance. However, they show large improvements over basic baselines on all dimensions while suggesting that a semantic understanding of arguments is required for the task.

Altogether, this paper's main contributions are:%
\footnote{The corpus and experiment code can be found under: \url{https://github.com/webis-de/ACL-23}}
\begin{itemize}
\setlength{\itemsep}{0ex}
\item
A theory-based taxonomy that specifies inappropriate language in online discussions
\item
A corpus with 2191~arguments from three different genres, manually annotated for the 14 taxonomy dimensions
\item
Empirical insights into the relation of appropriateness to previously studied quality dimensions and into its computational predictability
\end{itemize}

\section{Related Work}
\label{sec:related-work}

The notion of appropriateness has been explored in several sub-disciplines of linguistics. In communicative competence research, \citet{hymes:1972} considered the knowledge about cultural norms as a requirement to produce appropriate speech, which is a central part of acquiring communicative competence. Defining sociolinguistics, \citet{ranney:1992} linked appropriateness to the notion of politeness that is required in various social settings. Later, \citet{schneider:2012} argued that appropriateness is a more salient notion than politeness as it explicitly accounts for the context. Some of these cultural speech properties were identified as linguistic etiquette by \citet{jdetawy:2020}, including correct, accurate, logical, and pure language. 

Regarding the discussion of controversial issues, debating culture has required participants since its origins to stay on topic and to avoid offensive and overly emotional formulations \cite{andrew:1996}. Likewise, \citet{blair:1999} differentiate between good and bad bias in argumentation, where the latter exhibits close-mindedness, distortion of the conversation, or an imbalance of pro and con arguments. Similarly, \citet{walton:1999} introduced the concept of dialectical bias, explicitly addressing the context in which an argument is judged to be appropriate. This perspective on argumentation is also described by \citet{burkett:2011} as ``[...] making appropriate choices in light of situation and audience.'' 

As a sub-dimension of argument quality, appropriateness was first studied in NLP by \citet{wachsmuth:2017a}, a significant inspiration for our work. The authors derived appropriateness as one of the rhetorical argument quality dimensions based on the work of Aristotle \cite{aristotle:2007}. While several of the quality dimensions they proposed were addressed explicitly in previous work, the appropriateness dimension has not been systematically assessed until now. \citet{wachsmuth:2017a} only provided a relatively shallow definition of appropriateness that requires a simultaneous assessment of three properties, namely the creation of \emph{credibility} and \emph{emotions} as well as \emph{proportionality} to the issue. In contrast, we model these properties individually (in addition to several other dimensions) to better understand what exactly impacts appropriateness. 

Computationally, only \citet{wachsmuth:2020} tried to predict appropriateness alongside all the other quality dimensions of \citet{wachsmuth:2017a}. However, their models relied on a rather small sample of 304 arguments. In comparison, our corpus consists of~2191 arguments spanning three argumentative genres, providing deeper insights into the appropriateness of an argument. Related to this notion is the convincingness of arguments studied by \citet{habernal:2016b,habernal:2016a} which correlates with appropriateness \cite{wachsmuth:2017b}, as well as the effectiveness of arguments \cite{ng:2020,lauscher:2020}.

In the context of appropriateness, \citet{walton:2010} explored the notion of emotional fallacies in reasoning, some of which were later assessed computationally \cite{habernal:2017,alhindi:2022,jin:2022, goffredo:2022}.
Although we consider some of these fallacies in our work, we also consider other dimensions and exclude some irrelevant to appropriateness (i.e., logical fallacies) because of their more technical nature.

\bsfigure{dagstuhl-corpus-sets}{Venn diagrams showing the absolute counts of low-quality arguments in the corpus of \citet{wachsmuth:2017a} in terms of appropriateness and other dimensions: (a) The sub-dimensions of rhetorical effectiveness. (b) Local acceptability and global acceptability.}

We model \emph{toxic emotions} based on the emotional fallacies identified by \citet{walton:2010}: ad populum, ad misericordiam, ad baculum, and ad hominem. We merged these four into a single sub-dimension called \emph{emotional deception} based on the results of a pilot annotation study (Section \ref{sec:corpus}). Additionally, we define a sub-dimension \emph{excessive intensity} to address overly intense emotions. In particular, our analysis revealed the presence of a subset of propaganda errors, including loaded language, flag-waving, repetition, exaggeration, and minimization \citet{martino:2020}. 

\section{Modeling Appropriateness}
\label{sec:modeling}

\hwfigure{appropriateness-taxonomy-vertical}{Proposed taxonomy of inappropriate language in argumentation, with 14 dimensions and sub-dimensions. The colors are aligned with the argument quality dimensions used to derive them (Figure \ref{dagstuhl-corpus-sets}).}

This section explains how we established the relevant dimensions of appropriateness by systematically analyzing research on argument quality. 

\subsection{Appropriateness and Argument Quality}

To learn what makes an argument \emph{(in)appropriate}, we analyzed the interaction of appropriateness with other quality dimensions in the 304 arguments of \citet{wachsmuth:2017a}. We selected the dimensions that correlated most with appropriateness according to Pearson's $r$. These include the four sub-dimensions of rhetorical effectiveness (besides appropriateness), namely, \emph{credibility} ($.49$), \emph{emotional appeal} ($.30$), \emph{clarity} ($.45$), and \emph{arrangement} ($.48$), as well as \emph{local acceptability} ($.54$) (sub-dimension of logical cogency) and \emph{global acceptability} ($.59$) (sub-dimension of dialectical reasonableness). We then counted the number of arguments with the lowest quality rating for both appropriateness and the other dimensions as we expected the most notable differences in those instances. 

Figure \ref{dagstuhl-corpus-sets} illustrates the absolute cooccurrence of flawed arguments for the selected dimensions. Uniquely, appropriateness flaws always occur with at least one other flawed rhetorical dimension in all 43 cases, and low acceptability in nearly all cases.

Consequently, we manually analyzed arguments by contrasting pairs of arguments with and without low appropriateness to find patterns that describe what drives the low appropriateness levels within these dimensions. For example, to model the overlap of appropriateness with credibility, we compared the 29 arguments with only low credibility in Figure \ref{dagstuhl-corpus-sets} (a) to the 39 ($= 2+1+6+14+7+9$) arguments with low appropriateness and credibility. Concretely, we compared them incrementally, starting from arguments that do not have low values in any quality dimension except appropriateness and credibility, proceeding to those with exactly one other low value, and so forth until we reach the 14 arguments that have low values in all dimensions.

\subsection{Defining Inappropriateness}

The findings from our analysis led to four core inappropriateness dimensions in our taxonomy: We deem an argument \emph{inappropriate} (in light of its discussion context) if it is \emph{missing commitment} of its author to the discussion, uses \emph{toxic emotions}, is \emph{missing intelligibility}, or seems inappropriate for \emph{other reasons}. We detailed each in the following:

\paragraph{Toxic Emotions}

We model \emph{toxic emotions} based on the emotional fallacies identified by \citet{walton:2010}: ad populum, ad misericordiam, ad baculum, and ad hominem. We merged these four into a single sub-dimension called \emph{emotional deception} based on the results of a pilot annotation study (Section \ref{sec:corpus}). Additionally, we define a sub-dimension \emph{excessive intensity} to address overly intense emotions. In particular, our analysis revealed the presence of a subset of propaganda errors, including loaded language, flag-waving, repetition, exaggeration, and minimization \citet{martino:2020}. 

\paragraph{Missing Commitment}

This dimension resembles the \emph{credibility} dimension of \citet{wachsmuth:2017a}, but it differs in that we do not mandate arguments to come from or include a trusted source. 
Rather, the arguments should demonstrate the participant's general interest in participating in the debate. To formalize this concept, we drew on the five rules for ``A Good Dialogue'' \cite{walton:1999} to create two sub-dimensions of commitment, \emph{missing seriousness} and \emph{missing openness}, by examining the extent to which they apply to the arguments identified in the overlap analysis.

\paragraph{Missing Intelligibility} 

The core dimension \emph{missing intelligibility} results from the overlap analysis of the \emph{clarity} and \emph{arrangement} dimensions of \citet{wachsmuth:2017a}. We found that the main point of an argument was partly unclear either due to (un)intentional vagueness or overly (un)complex language, which we refer to in our taxonomy as the sub-dimension \emph{unclear meaning}. Also, derailing a discussion to another issue is a common issue (represented by the sub-dimension \emph{missing relevance}).
Finally, in some cases the individual claims and premises were intelligible but not their connection. We refer to this as a \emph{confusing reasoning}.

\paragraph{Other Reasons} 

This dimension accounts for reasons that do not fit into the other core-dimensions. As part of this, we observed that some arguments have a \emph{detrimental orthography}, limiting intelligibility in some cases (spelling or grammatical errors) or increasing emotions in others (capital letters, repeated exclamation points). We leave any other case of inappropriateness as \emph{reason unclassified}.

Figure \ref{appropriateness-taxonomy-vertical} depicts the final taxonomy of all 14 dimensions we propose. We hierarchically decompose \emph{inappropriateness} into the four core dimensions and those further into the nine discussed sub-dimensions to obtain a nuanced understanding of inappropriateness. The argument-centric focus of our taxonomy allows annotators to quickly formulate reasons for inappropriateness in the form ``$a$ is inappropriate because of $\sigma$'', where $a$ is an argument and $\sigma$ a specific sub-dimension from the taxonomy. We define each dimension below.

\subsection{A Hierarchical Taxonomy}

Since \emph{appropriateness} itself is already discussed in the literature, we refrain from redefining it here. Instead, we build on \citet{wachsmuth:2017a} who state that an argument ``has an appropriate style if the used language supports the creation of credibility and emotions as well as if it is proportional to the issue.'' Their annotation guidelines further suggest that ``the choice of words and the grammatical complexity should [...] appear suitable for the topic discussed within the given setting [...], matching the way credibility and emotions are created [...]''.

While our goal is to model appropriate language in argumentation, we decided to define when an argument is \emph{not} appropriate (as indicated above) to maintain freedom of speech as much as possible. Therefore, we define the four core dimensions and their sub-dimensions from Figure~\ref{appropriateness-taxonomy-vertical} in a ``reverse'' way, clarifying what is considered \emph{in}appropriate:

\paragraph{Toxic Emotions (TE)} 

An argument has toxic emotions if the emotions appealed to are deceptive or their intensities do not provide room for critical evaluation of the issue by the reader.\,\,
\begin{itemize}
\setlength{\itemsep}{0ex}
\item {\em Excessive Intensity (EI).} The emotions appealed to by an argument are unnecessarily strong for the discussed issue.
\item {\em Emotional Deception (ED).} The emotions appealed to are used as deceptive tricks to win, derail, or end the discussion. 
\end{itemize}

\paragraph{Missing Commitment (MC)}

An argument is missing commitment if the issue is not taken seriously or openness other's arguments is absent.
\begin{itemize}
\setlength{\itemsep}{0ex}
\item 
{\em Missing Seriousness (MS).} The argument is either trolling others by suggesting (explicitly or implicitly) that the issue is not worthy of being discussed or does not contribute meaningfully to the discussion.
\item 
{\em Missing Openness (MO).} The argument displays an unwillingness to consider arguments with opposing viewpoints and does not assess the arguments on their merits but simply rejects them out of hand. 
\end{itemize}

\paragraph{Missing Intelligibility (MI)}

An argument is not intelligible if its meaning is unclear or irrelevant to the issue or if its reasoning is not understandable.
\begin{itemize}
\setlength{\itemsep}{0ex}
\item 
{\em Unclear Meaning (UM).} The argument's content is vague, ambiguous, or implicit, such that it remains unclear what is being said about the issue (it could also be an unrelated issue).
\item 
{\em Missing Relevance (MR).} The argument does not discuss the issue, but derails the discussion implicitly towards a related issue or shifts completely towards a different issue. 
\item 
{\em Confusing Reasoning (CR).} The argument's components (claims and premises) seem not to be connected logically.
\end{itemize}

\paragraph{Other Reasons (OR)}

An argument is inappropriate if it contains severe orthographic errors or for reasons not covered by any other dimension.
\begin{itemize}
\setlength{\itemsep}{0ex}
\item 
{\em Detrimental Orthography (DO).} The argument has serious spelling and/or grammatical errors, negatively affecting its readability.
\item 
{\em Reason Unclassified (RU).} There are any other reasons than those above for why the argument should be considered inappropriate.
\end{itemize}

\section{The Appropriateness Corpus}  
\label{sec:corpus}

This section details the data acquisition and annotation process of our \emph{Appropriateness Corpus} and provides statistics of the collected annotations. Statistics of our corpus split by argument source are found in Appendix~\ref{sec:corpusstatistics}.


\begin{table*}[t!]
\small
\renewcommand{\arraystretch}{0.9}
\centering
\setlength{\tabcolsep}{0.75pt}
\begin{tabular*}{\linewidth}{@{}	l@{\;\;}l@{\;\;}r@{\;\;}r r@{\quad} c@{\;\;}c	r@{\quad\;}	r@{\quad}rrr@{\quad}rrr@{\quad}rrrr@{\quad}rrr}
\toprule
		&				& \multicolumn{2}{c}{\bf (a) Count} && \multicolumn{2}{c}{\bf (b) Agree.}			&& \multicolumn{14}{c}{\bf (c) Kendall's $\tau$ Correlation} \\
						   \cmidrule(l@{-2pt}r@{-2pt}){3-4}	\cmidrule(l@{-2pt}r@{-2pt}){6-7}	      \cmidrule(l@{-2pt}r@{0pt}){9-22}	
    \multicolumn{2}{l}{\bf Dimension} 	& \bf Yes  & \bf No && \phantom{.} \bf Full & \phantom{.}\bf $\alpha$ && \bf In	& \bf TE	& \bf EI	& \bf ED	& \bf MC	& \bf MS	& \bf MO	& \bf MI	& \bf UM	& \bf MR	& \bf CR	& \bf OR	& \bf DO	& \bf RU\\
\midrule		
\bf In	& \bf Inappropriateness   	& \bf 1182  &  1009	 && \phantom{..}60\tiny\%	  &.45        &&   	        & .56		& .38		& .44		& .59  	    & .35		& .47		& .62		& .41		& .42  	    & .25  	    & .21		& .18		& .10    \\[1ex]%
\bf TE	& \bf Toxic Emotions    	    	& 594       &  1597	     && \phantom{..}77\tiny\%	  &.36        && .56	    & 	        & \bf .66	& \bf .78  	& .35  	    & .11		& .35		& .13		& .01		& .12		& .06		& .00		& .00  	    & .00    \\
EI		& Excessive Intensity   	& 402       &  1789	     && \phantom{..}82\tiny\%     &.27        && .38	    & .66	    &		    & .22		& .25		& .06   	& .26		& .11		& .00		& .09		& .07		& .00		& .00		& .01    \\
ED		& Emotional Deception   		& 427       &  1764	     && \phantom{..}82\tiny\%	  &.36        && .44	    & \bf.78    & .22	    &     	    & .28		& .11		& .25		& .09		& .01		& .09		& .03		& .01		& .00		& .00    \\[1ex]%
\bf MC	& \bf Missing Commitment        	& 735       &  1456	     && \phantom{..}69\tiny\%	  &.21        && .59	    & .35		& .25		& .28		&           & \bf .57	& \bf .81	& .22		& .08		& .20		& .07		& .02		& .02		& .01    \\
MS		& Missing Seriousness     & 183       &  2008	     && \phantom{..}93\tiny\%	  &.51        && .35	    & .11		& .06		& .11		& .57		&           & .12   	& .15		& .09		& .17		& .03		& .02		& .02		& .00    \\
MO		& Missing Openness	    & 658       &  1533	     && \phantom{..}71\tiny\%	  &.11        && .47	    & .35		& .26		& .25		& \bf .81   & .12		&           & .18		& .05		& .16		& .07		& .01		& .00		& .01    \\[1ex]%
\bf MI	& \bf Missing Intelligibility    	& 774  		&  1417	     && \phantom{..}68\tiny\%	  &.25        && \bf .62	& .13		& .11		& .09		& .22		& .15		& .18		& 	        & \bf .64	& \bf .66	& \bf .41	& .14   	& .15		& .02    \\
UM		& Unclear Meaning 		& 459       &  1732	     && \phantom{..}80\tiny\%	  &.16        && .41	    & .01   	& .00		& .01		& .08		& .09		& .05		& .64		&   	    & .17		& .21	    & .16       & .20		& 0.1    \\
MR		& Missing Relevance 	& 508       &  1683	     && \phantom{..}78\tiny\%	  &.19        && .42	    & .12		& .09  	    & .09		& .20		& .17		& .16		& \bf .66	& .17		&		    & .07		& .03		& .02	    & .02    \\
CR		& Confusing Reasoning	& 174	    &  2017	     && \phantom{..}92\tiny\%     &.16     	  && .25	    & .06		& .07		& .03		& .07		& .03		& .07		& .41		& .21		& .07		& 	        & .14		& .15		& .01    \\[1ex]%
\bf OR	& \bf Other Reasons   	        	& 108       &  2083	     && \phantom{..}95\tiny\%	  &.23        && .21	    & .00		& .00		& .01		& .02	    & .02		& .01		& .14		& .16		& .03		& .14		& 	        & \bf .88	& \bf.43 \\
DO		& Detrimental Orthography    	    	& 77        &  2114	     && \phantom{..}97\tiny\%	  &.31        && .18	    & .00		& .00		& .00		& .02		& .02	    & .00   	& .15		& .20		& .02		& .15		& \bf .88	& 	        & .01    \\
RU		& Reason Unclassified	        & 32        &  \bf 2159  && \phantom{..}\bf 99\tiny\% &.00 	      && .10	    & .00		& .01		& .00		& .01		& .00		& .01	    & .02		& .01		& .02		& .01		& .43		& .01		&        \\
\bottomrule
\end{tabular*}
\caption{Corpus statistics of the 2191 annotated arguments: (a) Counts of annotations for each inappropriateness dimension, when being aggregated conservatively (i.e., at least one annotator chose \emph{yes}). (b) \emph{Full} agreement and Krippendorff's $\alpha$ agreement of all three annotators. (c) Kendall's $\tau$ correlation between the 14 inappropriateness dimensions, averaged over the correlations of all annotators. The highest value in each column is marked in bold.}
\label{table-agreement-and-correlations}
\end{table*}

\subsection{Data Acquisition}

Studying the applicability of our taxonomy requires a set of arguments that is both diverse and sufficiently large. We rely on manually labeled examples of reasonable quality to ensure that our corpus only contains argumentative texts. In particular, we collected all 2191 arguments on~1154 unique issues from existing corpora \cite{habernal:2016a,wachsmuth:2017a,ng:2020}.%
\footnote{The arguments from \newcite{wachsmuth:2017a} are a subset of those from \newcite{habernal:2016a} with additional annotations. We include each argument once only.}
All corpora are used in research on argument quality assessment \cite{habernal:2016b,wachsmuth:2020,lauscher:2020} and contain annotations that we identified as related to appropriateness: 
\begin{itemize}
\setlength{\itemsep}{0ex}
\item The Dagstuhl-15512 ArgQuality corpus \cite{wachsmuth:2017a} covers appropriateness and its most correlated dimensions.
\item The UKPConvArg2 \cite{habernal:2016b} corpus has reason labels for why argument $a$ is more convincing than argument~$b$.
\item The GAQCorpus \cite{ng:2020} covers four argument quality dimensions, including effectiveness, the ``parent'' of appropriateness.
\end{itemize}

We carefully selected the source corpora such that about $50\%$ of the arguments belong to only 16 issues while the rest covers the remaining 1138 issues, making our corpus valuable both vertically (issues with many arguments allow deeper analyses) and horizontally (large number of issues promotes generalizability). The average sentence length of arguments is 4.8. The corpus includes arguments of three genres, 1590 from debate portals, 500 from question answering forums, and 101 reviews.

\subsection{Annotation Process}

We designed a task-specific annotation interface that leverages the hierarchical structure of the taxonomy in Figure \ref{appropriateness-taxonomy-vertical}. Specifically, annotators needed to label sub-dimensions, only if the respective core dimension was labeled before as given for an argument. Following \citet{wachsmuth:2017a}, we used an ordinal scale for the \emph{inappropriateness} dimension described as (1) fully inappropriate, (2) partially (in)appropriate, and (3) fully appropriate. 

Likewise, a binary yes/no scale was used for all the other dimensions, where \emph{yes} means inappropriateness in terms of the respective dimension.
Annotators were required to select a reason (core dimension) from the taxonomy only for partially or fully inappropriate arguments. We provided a coherent and self-descriptive interface (see Appendix~\ref{sec:annotationinterface}) to reduce the cognitive load on the annotators. The annotators also had the opportunity to provide their own reasons for the \emph{reason unclassified} dimension. 

We conducted two rounds of annotation to find qualified annotators. In the first round, eight native English speakers hired on \emph{Upwork} and two authors of this paper (5 female, 5 male in total) each annotated 100 arguments, randomly sampled from our corpus. Based on the results and feedback on the annotation interface and the guidelines, we refined our taxonomy, most notably reducing the number of dimensions from 18 to 14. For the second round, we selected the three Upwork annotators with the highest expert correlations (2 female, 1 male). We paid \$13 per hour for annotating all 2191 arguments, as we did in the first round. To mitigate the cognitive overload entailed by prolonged reading, we divided the annotation into 14 batches of roughly 150 arguments each and limited the number of batches to be annotated per day to one.

\subsection{Corpus Statistics and Agreement}

\begin{table}[t!]
\small
\centering
\setlength{\tabcolsep}{3.5pt}
\begin{tabular*}{\linewidth}{l@{}rrr}
\toprule
&				\multicolumn{3}{c}{\bf Krippendorff's $\alpha$}  \\
			    \cmidrule(l@{-2pt}r@{-2pt}){2-4}		
 \textbf{Dimension} & \textbf{Liberal} & \textbf{Majority} & \textbf{Conservat.} \\ \midrule
\bf Inappropriateness & 0.16 & 0.54 & \textbf{0.95} \\[1ex]
\bf Toxic Emotions & 0.14 & 0.45 & \textbf{1.00} \\
Excessive Intensity & -0.08 & 0.30 & \textbf{1.00} \\
Emotional Deception & 0.05 & 0.41 & \textbf{1.00} \\[1ex]
\bf Missing Commitment & -0.03 & 0.30 & \textbf{1.00} \\
Missing Seriousness & 0.27 & 0.54 & \textbf{1.00} \\
Missing Openness & -0.12 & 0.22 & \textbf{0.96} \\[1ex]
\bf Missing Intelligibility & -0.03 & 0.41 & \textbf{1.00} \\
Unclear Meaning & -0.07 & 0.19 & \textbf{1.00} \\
Missing Relevance & -0.04 & 0.22 & \textbf{1.00} \\
Confusing Reasoning & -0.04 & 0.19 & \textbf{0.95} \\[1ex]
\bf Other Reasons & 0.08 & 0.31 & \textbf{1.00} \\
Detrimental Orthography & 0.13 & 0.42 & \textbf{1.00} \\
Reason Unclassified & -0.01 & -0.01 & \textbf{1.00} \\ \bottomrule
\end{tabular*}
\caption{Krippendorff's $\alpha$ agreement between MACE labels and the manual labels obtained by each evaluated combination strategy (liberal, majority,  conservative).}
\label{table-mace}
\end{table}

To combine the annotators' labels in our corpus, we first use MACE \cite{hovy:2013} in order to consider the annotators' reliability. We then compute Krippendorff's $\alpha$ between the MACE labels and those obtained with either of three combination strategies: \emph{Liberal} considers an argument appropriate if at least one annotator marked it as such. \emph{Majority} considers the label for which at least two annotators agree. \emph{Conservative}, finally, considers an argument inappropriate if at least one annotator marked it as such. Table~\ref{table-mace} shows that the MACE labels correlate best with the conservative labels in all cases. Consequently, to obtain the final corpus annotations, we combined the three labels of each argument following the conservative strategy. This strategy also seems most consistent with the current belief system in many societies around the world, that is, to accommodate minorities in language.

\begin{table*}[t!]
\small
\renewcommand{\arraystretch}{0.9}
\centering
\setlength{\tabcolsep}{0.65pt}
\begin{tabular*}{\linewidth}{ll@{}rr@{\quad}rrr@{\quad}rrr@{\quad}rrrr@{\quad}rrrr}
\toprule
		&				&  \multicolumn{15}{c}{\bf Kendall's $\tau$ Correlation with Inappropriateness Dimensions} \\
						   \cmidrule(l@{-5pt}r@{0pt}){4-17}	
\bf Quality Dimension	&  \bf Description		&&  \bf In	& \bf TE	& \bf EI	& \bf ED	& \bf MC	& \bf MS	& \bf MO	& \bf MI	& \bf UM	& \bf MR	& \bf CR	& \bf OR	& \bf DO	& \bf RU\\
\midrule
\bf Cogency    			& $P$ acceptable / relevant / sufficient				&& .27      & .18       & .19	    & .13	    & .22	    & .22	    & .15	    & .27	    & .20	    & .21	    & .20	    & .12	    & .11	    & \bf .14	\\
Local acceptability   	& $P$ rationally believable								&& .36      & \bf .29	& \bf .27   & \bf .26	& .28	    & .23	    & .21	    & .32	    & .20	    & .26	    & .23	    & .13	    & .12	    & .07	    \\
Local relevance    		& $P$ contribute to acceptance / rejection				&& .30      & .16	    & .16	    & .14       & .22	    & .26	    & .13	    & .31	    & .21	    & .25	    & .21	    & .13	    & .14       & .06	    \\
Local sufficiency    	& $P$ give enough support								&& .25      & .15	    & .15	    & .10	    & .20       & .19	    & .14	    & .27	    & .20	    & .22	    & .17	    & .10	    & .07	    & .07	    \\[1ex]%
\bf Effectiveness     	& $a$ persuades target audience						    && .27      & .18	    & .18	    & .13	    & .23	    & .21       & .17	    & .26	    & .19	    & .20	    & .21	    & .12	    & .11	    & .04	    \\
Credibility    			& $a$ makes author worthy of credence					&& .32      & .22	    & .16	    & .19	    & .29	    & .24	    & .22       & .29	    & .22	    & .21	    & \bf .25	& .15	    & .12	    & .07	    \\
Emotional appeal	    & $a$ makes target audience more open					&& .17      & .13	    & .10       & .13	    & .16	    & .15	    & .12	    & .14       & .15	    & .14	    & .16	    & .10	    & .04	    & .08	    \\
Clarity 		    	& $a$ uses correct/unambiguous language                 && .20      & .09	    & .08	    & .07	    & .10	    & .19	    & .04	    & .25	    & .18       & .22	    & .21	    & .14	    & \bf .21	& .08	    \\
Appropriateness			& $a$'s credibility/emotions are proportional  			&& \bf .41  & .25	    & .24	    & .21	    & \bf .35	& \bf .28	& \bf .27	& \bf .36	& \bf .30	& .25       & .21	    & \bf .20	& .20	    & .08	    \\	
Arrangement    			& $a$ has components in the right order					&& .26      & .13	    & .15	    & .10	    & .16	    & .19	    & .10	    & .31	    & .25	    & .21	    & .24       & .13	    & .15	    & .02	    \\[1ex]%
\bf Reasonableness    	& $A$ acceptable / relevant / sufficient				&& .34      & .23	    & .23	    & .16	    & .27	    & .23	    & .20	    & .32	    & .24	    & .26	    & .20       & .16       & .14	    & .11	    \\
Global acceptability 	& $A$ worthy to be considered							&& .38      & .28	    & \bf .27	& .22	    & .31	    & .25	    & .24	    & .35	    & .24	    & \bf .27	& \bf .25	& .16	    & .16       & .07	    \\
Global relevance 		& $A$ contribute to issue resolution				    && .24      & .16	    & .21	    & .11	    & .17	    & .23	    & .10	    & .26	    & .19	    & .21	    & .18	    & .12	    & .11	    & .08	    \\
Global sufficiency		& $A$ adequately rebuts counterarguments 				&& .20	    & .15       & .19	    & .09	    & .21	    & .16	    & .16	    & .19	    & .17	    & .12	    & .15	    & .06	    & .04	    & .08	    \\[1ex]%
\bf Overall quality    	& $a/A$ is of high quality								&& .30      & .19	    & .20	    & .15	    & .24	    & .22	    & .17	    & .30	    & .24	    & .21	    & .21	    & .14	    & .12	    & .10	    \\

\bottomrule 
\end{tabular*}
\caption{Kendall's $\tau$ correlation of the mean ratings of the argument \emph{quality dimensions} of \citet{wachsmuth:2017a} with the 14 proposed \emph{inappropriateness dimensions} (see Table~\ref{table-agreement-and-correlations} for the meaning of the acronyms). $P$ are the premises of an argument $a$ that is used within argumentation $A$. The highest value in each column is marked in bold.}
\label{table-correlations-dagstuhl}
\end{table*}


Table \ref{table-agreement-and-correlations}(a) presents the corpus distribution of the annotations aggregated conservatively. For readability, we binarized the overall inappropriateness in the table, considering both fully and partially inappropriate arguments as inappropriate. 1182 arguments were considered at least partially (in)appropriate (540 of them fully inappropriate). 

Among the reasons given, \emph{missing intelligibility} is the most frequent core dimension (774 arguments) and \emph{missing openness} the most frequent sub-dimension (658), matching the intuition that a missing openness to others' opinions is a key problem in online discussions. The least frequent core dimension is \emph{other reasons} (108), and the least frequent sub-dimension \emph{reason unclassified} (32). That is, our annotators rarely saw additional reasons, indicating the completeness of our taxonomy. 

Table \ref{table-agreement-and-correlations}(b) shows inter-annotator agreement. For \emph{inappropriateness}, the annotators had full agreement in 60\% of all cases, suggesting that stricter settings than our conservative strategy can also be applied without limiting the number of annotations too much. The Krippendorff's $\alpha$ agreement is limited but reasonable given the subjectiveness of the task. It ranges from $.11$ to $.51$ among the dimensions (not considering \emph{reason unclassified}), with $.45$ for overall \emph{inappropriateness}. These values are similar to those of \citet{wachsmuth:2017a}.

\begin{table*}[t!]
\small
\renewcommand{\arraystretch}{0.9}
\centering
\setlength{\tabcolsep}{0.95pt}
\begin{tabular*}{\linewidth}{ll r @{\;\;}r@{\;\;\;}rrr@{\;\;\;}rrr@{\;\;\;}rrrr@{\;\;\;}rrr}
\toprule
		&				&  \multicolumn{15}{c}{\bf Kendall's $\tau$ Correlation with Inappropriateness Dimensions} \\
						   \cmidrule(l@{-5pt}r@{0pt}){4-17}	
\bf Comparison		&  \bf Reason		&&  \bf In	& \bf TE	& \bf EI	& \bf ED	& \bf MC	& \bf MS	& \bf MO	& \bf MI	& \bf UM	& \bf MR	& \bf CR	& \bf OR	& \bf DO	& \bf RU\\
\midrule
\multirow{3}{1.91cm}[-0em]{$b$ is less convincing than $a$, since...}
				& $b$ is attacking / abusive			                      			    && \bf .86	& \bf .70	& \bf .54	& \bf .70	& \bf .70		& .65	    & \bf .44	& .49	    & .34	    & .50	    & .01		    & .05	    & .02	    & .13		    \\
				& $b$ has language issues / humour / sarcasm			                    && .77		& .35	    & .21	    & .35	    & .61		    & \bf .69	& .14	    & .44	    & .41	    & .35	    & .16		    & .34	    & .33	    & .15		    \\
				& $b$ is unclear / hard to follow			                  			    && .61		& .16	    & .03	    & .17	    & .30		    & .36	    & .13	    & .52	    & .54	    & .33	    & \bf .36	    & \bf .48	& \bf .47	& .11		    \\
				& $b$ has no credible evidence / no facts	                  			    && .56		& .20	    & .20	    & .12	    & .37		    & .46	    & .21	    & .47	    & .41	    & .29	    & .24		    & .15	    & .11	    & .12		    \\
				& $b$ has less or insufficient reasoning	                  			    && .82		& .32	    & .29	    & .24	    & .59		    & .64	    & .34	    & .63	    & \bf .57	& .46	    & .25		    & .08	    & .04	    & .16		    \\
				& $b$ uses irrelevant reasons 				                  			    && .69	    & .29	    & .21	    & .25	    & .45	        & .49	    & .25	    & .60	    & .44	    & .54	    & .20	        & .20	    & .09	    & .02	        \\
				& $b$ is only an opinion / a rant			                  			    && .72		& .21	    & .18	    & .19	    & .45		    & .55	    & .22	    & .57	    & .42	    & .44	    & .16		    & .19	    & .09	    & .14		    \\
				& $b$ is non-sense / confusing				                  			    && .69		& .22	    & .09	    & .25	    & .48		    & .57	    & .14	    & .58	    & .49	    & .43	    & .32		    & .40	    & .37	    & .06		    \\
				& $b$ does not address the topic			                  			    && .82	    & .04	    & .03	    & .07	    & .30	        & .57	    & .09	    & \bf .75	& .48	    & \bf .72	& .05	        & .19	    & .00	    & .10	        \\
				& $b$ is generally weak / vague			                  			        && .59		& .18	    & .19	    & .11	    & .35		    & .42	    & .18	    & .53	    & .48	    & .33	    & .28		    & .08	    & .07	    & .15		    \\[1ex]
\multirow{3}{1.91cm}[-0em]{$a$ is more convincing than $b$, since...}                                                   
				& $a$ is more detailed / better reasoned / deeper                           && .50		& .14	    & .11	    & .12	    & .32		    & .42	    & .16	    & .45	    & .40	    & .30	    & .21		    & .19	    & .15	    & .11		    \\
				& $a$ is objective / discusses other views                               	&& .44		& .15	    & .09	    & .14	    & .30		    & .39	    & .18	    & .37	    & .36	    & .25	    & .17		    & .21	    & .15	    & .18		    \\
				& $a$ is more credible / confident			                              	&& .40		& .13	    & .05	    & .18	    & .20		    & .34	    & .10	    & .42	    & .38	    & .30	    & .18		    & .17	    & .12	    & .20		    \\
				& $a$ is clear / crisp / well-written		                              	&& .55	    & .30	    & .22	    & .27	    & .35	        & .37	    & .24	    & .47	    & .41	    & .31	    & .32		    & .27	    & .25	    & .20		    \\
				& $a$ sticks to the topic					                              	&& .65		& .23	    & .17	    & .22	    & .34		    & .49	    & .07	    & .55	    & .39	    & .47	    & .15	        & .19	    & .13	    & .18		    \\
				& $a$ makes you think 					                                  	&& .38		& .13	    & .15	    & .08	    & .22		    & .34	    & .07	    & .27	    & .19	    & .20	    & .21		    & .11	    & .14	    & \bf .26	    \\
				& $a$ is well thought through / smart		                              	&& .56	    & .26	    & .14	    & .27	    & .36	        & .40	    & .23	    & .46	    & .34	    & .27	    & .30	        & .28	    & .24	    & .05	        \\[1ex]

Overall   & $a$ is more convincing than $b$                                                 && .53      & .17	    & .13	    & .15	    & .32	        & .43	    & .14	    & .44	    & .37	    & .32	    & .19	        & .19	    & .14	    & .13	        \\
\bottomrule
\end{tabular*}
\caption{Kendall's $\tau$ correlation of the convincingness \emph{comparison reasons} of argument pairs $(a, b)$ of \citet{habernal:2016b} with differences in the mean ratings of dimensions of the proposed \emph{inappropriateness dimensions} (see Table~\ref{table-agreement-and-correlations} for the meaning of the acronyms). The highest value in each column is marked in bold.}
\label{table-correlations-ukp}
\end{table*}

\section{Analysis}
\label{sec:analysis}

Building on existing corpora on theoretical and practical argument quality, we now report the correlations of our proposed dimensions and the quality dimensions of \newcite{wachsmuth:2017a} and \newcite{habernal:2016b}. Correlations with \newcite{ng:2020} are found in Appendix~\ref{sec:gaqcorpus} (only one dimension is directly related to appropriateness).

\subsection{Relations between Corpus Dimensions}

Table \ref{table-agreement-and-correlations}(c) presents the Kendall's $\tau$ correlations between all inappropriateness dimensions. Among the core dimensions, we find \emph{missing intelligibility} to be most ($.62$) and \emph{other reasons} to be least ($.21$) correlated  with  \emph{inappropriateness (In)}. In case of the sub-dimensions, \emph{missing openness} is most ($.47$) and \emph{not classified} least ($.10$) correlated with it. 

The sub-dimensions are mostly correlated with their direct parent, with values between $.41$ and $.88$, which is expected due to our annotation study setup. However, there are clear differences between sub-dimensions of the same parent; for example, \emph{excessive intensity} and \emph{emotional deception} are highly correlated with \emph{toxic emotions} ($.66$ and $.78$) but have low correlation with each other ($.22$). Cross-dimensional correlations among the core- and sub-dimensions are highest between \emph{toxic emotions} and \emph{missing intelligibility} ($.35$) and \emph{excessive intensity} and \emph{missing openness} ($.28$) respecitvely.  This suggests that overly intense emotions sometimes signify a rejection of others' opinions and vice versa. 

\subsection{Relation to Theory of Argument Quality}

Table \ref{table-correlations-dagstuhl} shows the Kendall's $\tau$ correlations between our dimensions and the theoretical quality dimensions of \newcite{wachsmuth:2017a}. We observe the highest correlation for the two \emph{(in)appropriateness} dimensions ($.41$), showing that our annotation guideline indeed captures the intended information for the annotated arguments. Furthermore, seven of our dimensions correlate most strongly with \emph{appropriateness} in the Dagstuhl-15512 ArgQuality corpus, and all 14 dimensions have the highest correlation with one of the seven argument quality dimensions that we used to derive the taxonomy. 

The values of \emph{reason unclassified (RU)} are low (between $.02$ and $.14$), speaking for the completeness of our taxonomy. However, its most correlated quality dimension is \emph{cogency}, possibly indicating a minor logical component of appropriateness.

\subsection{Relation to Practice of Argument Quality}

Table \ref{table-correlations-ukp} shows the correlations between our dimensions and the convincingness comparison reasons of \citet{habernal:2016b}. We see that attacking/abusive behavior is most correlated \nopagebreak{with our} \emph{inappropriateness} (In, $.86$), \emph{missing commitment} (MC, $.70$) and \emph{toxic emotions} (TE, $.70$) dimensions. \emph{Missing seriousness (MS)} and \emph{missing intelligibility (MI)} are mostly correlated with humor/sarcasm ($.69$) and not addressing (derailing) the topic ($.75$) respecitvely. \emph{Confusing reasoning (CR)} is most correlated with an argument being hard to follow ($.36$), and \emph{unclear meaning (UM)} with insufficient reasoning ($.57$). 

\begin{table*}[t!]
\small
\renewcommand{\arraystretch}{.9}
\centering
\setlength{\tabcolsep}{3.5pt}
\begin{tabular*}{\linewidth}{	l	r	r@{\quad\;}rrr@{\quad\;}rrr@{\quad\;}rrrr@{\quad\;}rrr@{\quad\;}l}
\toprule
    \bf Approach  & \phantom{aa}  & \bf In	& \bf TE	& \bf EI	& \bf ED	& \bf MC	& \bf MS	& \bf MO	& \bf MI	& \bf UM	& \bf MR	& \bf CR	& \bf OR	& \bf DO	& \bf RU	& \bf Macro	\\
\midrule
Random baseline 						 &&	.49  	 &	.47  	 &	.45  	 &	.45  	 &	 .49  	  &	  .39  	   &	.47  	 &	 .48 		&	.45  	   &	.47 	 &	.39 	 &	.37  	 &	.37  	 &	.34 	&	\phantom{.} .43  	                    \\
Majority baseline 					 	 &&	.32  	 &	.42  	 &	.45  	 &	.45  	 &	 .40  	  &	  .48 	   &	.41  	 &	 .39 		&	.44  	   &	.43  	 &	.48  	 &	.49  	 &	.49  	 &	.50 	&	\phantom{.} .44  	                    \\
\addlinespace
DeBERTaV3-large                          && \bf .75 	& \bf .74 	& \bf .69 	& \bf .70 	& \bf .75 	& \bf .73 	& \bf .72 	& \bf .72 	& \bf .69 	& .68 		& \bf .62 	& \bf .65 	& \bf .67 	& \bf .52 	& \phantom{.} \bf .69$^{\dagger\ddagger}$          	\\
\quad DeBERTaV3-w/o-issue                      && \bf .75 	& .73 		& .68 		& \bf .70 	& \bf .75 	& \bf .73 	& .71 		& \bf .72 	& .68 		& \bf .69 	& .61 		& .63 		& .66 		& .51 		& \phantom{.} .68$^{\ddagger}$ 		                \\
\quad DeBERTaV3-shuffle                        && .72 		& .69 		& .64 		& .64 		& .71 		& .65	 	& .68 		& .70 		& .66 		& .65 		& .57 		& .59 		& .57 		& .50 	    & \phantom{.} .64		                            \\
\midrule
Human performance 						     &&	.78 	 &	.79  	 &	.73  	 &	.77  	 &	 .73  	  &	  .82  	   &	.70  	 &	 .76 		&	.73  	   &	.72  	 &	.74  	 &	.78  	 &	.80  	 &	.70 	&	\phantom{.} .75 	                    \\
\bottomrule 
\end{tabular*}
\caption{Evaluation of appropriateness classification: F$_1$-score of each approach in 5-times repeated 5-fold cross validation on all 14 proposed dimensions. The best value in each column is marked bold. We marked significant macro F$_1$-score gains over \emph{DeBERTaV3-w/o-issue} ($\dagger$) and \emph{DeBERTaV3-shuffle} ($\ddagger$) at $p < .05$.}
\label{table-scores}
\end{table*}

We find that \emph{detrimental orthography (DO)} renders an argument unclear and difficult to follow ($.47$). Finally, the \emph{reason unclassified (RU)} dimension is most correlated with making a reader think about an argument. Manual inspection of the reasons for these annotations reveals that annotators chose \emph{reason unclassified}, if they were unsure which of the other dimensions they should assign.

\section{Experiments}    
\label{sec:experiments}  

The corpus from Section~\ref{sec:corpus} is meant to enable the computational treatment of inappropriate language in argumentation. As an initial endeavor, this section reports baselines for classifying all 14 dimensions in the taxonomy from Section~\ref{sec:modeling}.

\subsection{Experimental Setup}

In line with Table~\ref{table-agreement-and-correlations}, we treat all annotations as binary labels. We performed five repetitions of 5-fold cross-validation (25 folds in total) and ensured a similar distribution of the labels in each fold. For each folding, we used 70\% for training, 10\% for selecting the best-performing approach in terms of the mean macro-F$_1$ score, and 20\% for testing.

\paragraph{Models}

For classification, we employed the recent model \emph{DeBERTaV3-large} \cite{he:2021}, with an argument prepended by the discussion issue as input. Besides, we tested two ``ablations'': \emph{DeBERTaV3-w/o-issue} receives only the argument to gain insight into how effective it is to provide the issue as context. \emph{DeBERTaV3-shuffle} receives the argument and the issue with all words shuffled, to analyze the impact of proper syntactic and semantic formulations. We trained our models to predict all 14 dimensions via a multi-label prediction loss, accounting for data imbalance by assigning weights to all dimensions (more details in Appendix \ref{sec:hyperparameters}).

\paragraph{Lower and Upper Bounds}

To quantify the impact of learning, we compare against a \emph{random baseline} that chooses a label pseudo-randomly and a \emph{majority baseline} that takes the majority label for each dimension. As an upper bound, we measure \emph{human performance} in terms of the average of each human annotator in isolation on the dataset.

\subsection{Results}

Table \ref{table-scores} presents the mean F$_1$-score for all 14 inappropriateness dimensions averaged over all folds. \emph{DeBERTaV3-large} performs best in terms of macro F$_1$-score ($.69$), significantly beating both \emph{DeBERTaV3-w/o-issue} ($.68$) and \emph{DeBERTaV3-shuffle} ($.65$) in a Wilcoxon signed-rank test ($p < .05$) . The gain over \emph{DeBERTaV3-w/o-issue} is small though, suggesting that the context of a discussion (here, the issue) may be of limited importance for predicting inappropriateness. Plausible reasons are that (1) most arguments are (in)appropriate regardless of their context, or (2) the context of the argument is explicitly or implicitly contained within most arguments. \emph{DeBERTaV3-w/-issue} clearly outperforms the random baseline and majority baseline on all dimensions, and it achieves about 92\% of human performance in terms of macro F$_1$ ($.75$). These results suggest the possibility of automating the task of predicting appropriateness, however, encouraging further improvements.




\section{Conclusion}
\label{sec:discussion-and-conclusion}

Online discussions of controversial topics mostly turn out fruitful only, when the participants argue \emph{appropriately}, a dimension of argumentative language that has received no systematic investigation so far. Therefore, we have presented a taxonomy of 14 dimensions to model inappropriate language in argumentation, derived from rhetoric and argumentation theory. To enable computational research on appropriateness, we compiled a corpus of 2191 arguments from three genres, carefully annotated for all dimensions. 

Our extensive corpus analyses confirm correlations with both theoretical and practical dimensions of argument quality from the literature. The taxonomy covers inappropriateness comprehensively according to human annotators. While a DeBERTa-based baseline already comes rather close to human performance in classifying inappropriate language, our corpus allows for developing more sophisticated models in future work that may serve an automatic (or semi-automatic) content moderation. 

To make content moderation successful and accepted, we think that providing clear reasons supporting the moderation is important, so the participants can better frame their arguments in online discussions. The defined taxonomy dimensions lay out how such reasons may look like.


\section{Acknowledgments}

This project has been partially funded by the German Research Foundation (DFG) within the project OASiS, project number 455913891, as part of the Priority Program ``Robust Argumentation Machines (RATIO)'' (SPP-1999). We would like to thank the participants of our study and the anonymous reviewers for the feedback and their time.

\section{Limitations}

Aside from the still-improvable performance of the classification models we evaluated, our work is limited in two ways:  the nature of what is considered appropriate as well as the difficulties that arise during corpus creation in NLP in general. 

We point to the subjectivity in perception regarding appropriateness, which is also displayed and discussed in the paper by the inter-annotator agreement. Many sociocultural factors can influence this perception within cultures, such as age, gender, education, or ethnicity. We sought to account at least for gender by including both male and female annotators for all arguments. However, we encourage further studies that focus on other factors, as we expect appropriateness to be seen differently, primarily across cultures with varying styles of debates. Since our corpus contains only arguments written in English and is annotated by native English speakers, it may also be insufficient to generalize across languages. 

Moreover, appropriateness perception is likely subject to change over time. Although we collected arguments from different years, we see long-time limitations to our corpus. In general, it also depends on the expectations of the discussion participants, which are to some extent predetermined by the context (e.g., a sales pitch vs.\ a discussion with friends). In that regard, the context of our corpus is solely that of discussing controversial issues with strangers on the web. Finally, the size of the created corpus we propose in the paper may limit the generalizability of approaches that build on it and should be investigated further in future work.

\section{Ethical Considerations}

The corpus and the computational baselines presented in this paper target a sensitive issue: what is considered appropriate to say in a discussion. We suggest differentiating between freedom of speech, hate speech, and inappropriate speech. We believe inappropriate speech is an extension of hate speech that leads to a less free but more healthy climate in speech exchange. While freedom of speech in many countries is limited by hate speech in law, the extension to inappropriate speech is not. Consequently, automating the detection of inappropriateness and dealing with it in the same way hate speech is addressed (often by removal) may be perceived as hurting individuals' freedom of speech and, thus, must be handled with great care. 

However, we see no strong immediate ethical concerns regarding the computational methods specific to our work, as they only detect inappropriateness and do not recommend any actions. We stress, though, that they are not meant yet for real-life applications. Apart from the outlined limitations, we also do not see notable ethical concerns regarding our taxonomy, as we derived it systematically from existing literature and always encouraged our annotators to add their own reasons.

Finally, we aimed to ensure fair payment. As discussed in the paper, our annotators were paid about \$13 per hour, which exceeds the minimum wage in most US states and is also conform to the standards in the regions of our host institutions.

\bibliography{acl23-appropriateness-corpus-lit}
\bibliographystyle{acl_natbib}
\appendix
\section{Training Hyperparameters}
\label{sec:hyperparameters}
We did a single initial round of hyperparameter optimization and sticked to the best values for all of our DeBERTaV3 experiments: a polynomial learning rate with a warmup ratio of $.10$, a batch size of $10$, and an initial learning rate of $3 \cdot 10^{-6}$, trained for 10 epochs in all cases.

\section{Computational Infrastructure}
Our experiments were done on Ubuntu 20.04 with Python version 3.7.12, CUDA version 11.3 and one A100-SXM4-40GB GPU.
We used the following main libraries in our experiments (we include a full list of packages and their versions in the requirements.txt in the supplementary material):
\begin{itemize}
    \item torch==1.10.2+cu113
    \item transformers==4.21.0.dev0
    \item spacy==3.3.1
\end{itemize}
\clearpage
\onecolumn
\section{Examples}
\label{sec:examples}
\newcolumntype{b}{X}
\newcolumntype{s}{>{\hsize=.5\hsize}X}

\begin{table*}[h!]
\small
\centering
\begin{tabularx}{\textwidth}{ssb}
    \toprule
\textbf{Type} & \textbf{Issue} & \textbf{Argument} \\
\midrule
\bf Toxic Intensity & If given the choice what way would you vote for the death penalty to be brought back in Britain? & Yes i am completely for it. People are arguing that it is barbaric and inhumane but who can stand up and say that some perv who has raped and killed a child still has human rights and the right to live. We would put down a dangerous dog why not do the same to some of the scum that lives in our country. The justice system in britain at the moment is hopeless. Far to many people are gettin away with all sorts and something needs to be done!! \\\midrule
\bf Emotional Deception & Can the police keep your car permantly if you have 3rd suspended license? & Towed three times and impounded for 30 days each time? Man, you're just not getting the message, are you?If you are in California, you bet the police can forfeit your vehicle and it doesn't take three times to make it a charm. Technically, your vehicle could be subject to forfeiture proceedings after your first suspended license beef. Someone like you is exactly the reason the legislature designed that law, because your privilege to drive has been taken away from you and yet you obviously continue to drive. People like you are involved in an exponentially higher than average number of traffic accidents so the legislature figured maybe people like you should have your vehicles forfeited to the state if you just didn't go along with the game plan.Voila - I give you California Vehicle Code section 14607.6...and a link to it below. It would also be worth your time to review 14607.4, whether or not you live in California.You really need to stop driving. Really. \\\midrule
\bf Missing Seriousness & is-porn-wrong & Porn is Wrong. mainly because they are Not Doing it Right. it should be Hi Def. in three years, it will be in 3-D. \\\midrule
\bf Missing Openness & Pro-choice-vs-pro-life & There should be no argument in this really...whatever way yu see a fetus...its still a living form that has been created in a very intimate way... you shouldn't be changing what mothernature or God or fate or whatever has decided for you...and if you didn;t wannna get preggo in the first place...don't have sex or use protection. Yeh there are some women that get raped and it's very unfortunate but they should give the child up for adoption. It's not the child's fault that it was created. So why should the goring being have to pay the ultimate price of it's life? \\\midrule
\bf Unclear Meaning & Evolution-vs-creation & Believing "Evolution" as in Darwinism and the like, is like believing the puzzle can be solved by pouring the pieces out because two pieces kind of stuck together. \\\midrule
\bf Missing Relevance & Is it illegal to record a phone conversation? & The conversation can not be used as evidence in a court of law. I don't know what the lady hoped to gain from recording the conversation other than to create more drama. Some people are hooked on drama and they actually do what they can to create it. Run as far away and as fast as you can from these types. They will suck you dry. \\\midrule
\bf Confusing Reasoning & If your spouse committed murder and he or she confided in you would you turn them in? & i would turn in my wife because its wrong to kill someone. it could have been an accident but it was still wrong and besides the police are going to find out who killed that person but i don't want her to leave me for a long period of time so i would tell but then again i wouldn't. \\\midrule
\bf Deceptive Orthography & Is-the-school-uniform-a-good-or-bad-idea & it dose not show kids expressions and unforms dose not show is it \\\midrule
\bf Reason Unclassified & Firefox-vs-internet-explorer & Firebug, WebDeveloper, TabMix, FaviconizeTab, GreaseMonkey, IETab (to use when you visit microsot.com). Just some reason why i prefer Firefox \\
\bottomrule
\end{tabularx}
\caption{Examples of inappropriate arguments from our corpus for each of the nine sub-dimensions of our taxonomy.}
    \label{table-examples}
\end{table*}

\clearpage

\section{Annotation Interface}
\label{sec:annotationinterface}

\begin{figure}[h!]
\includegraphics[width=450pt]{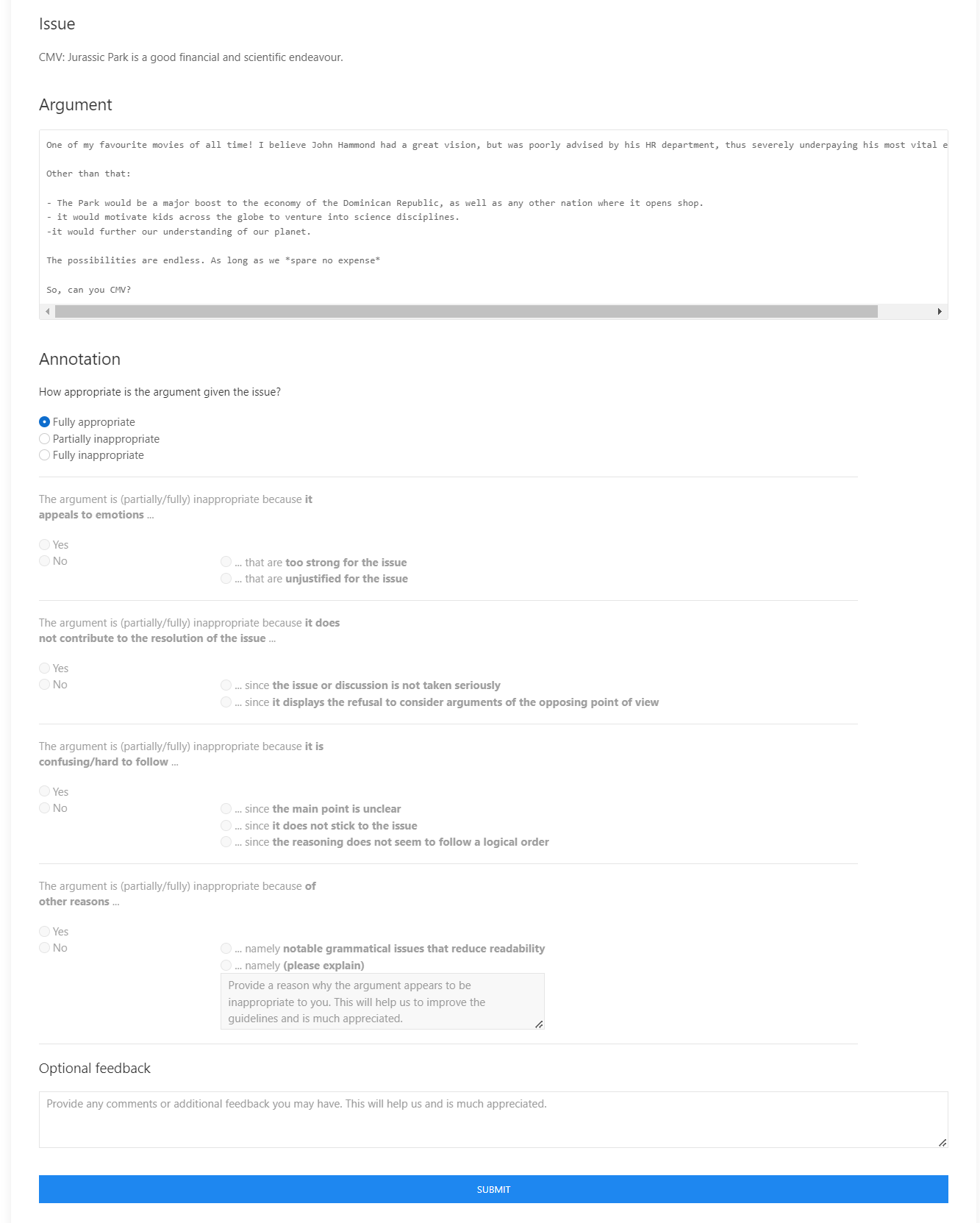}
\end{figure}

\clearpage

\begin{figure}[h!]
\includegraphics[width=450pt]{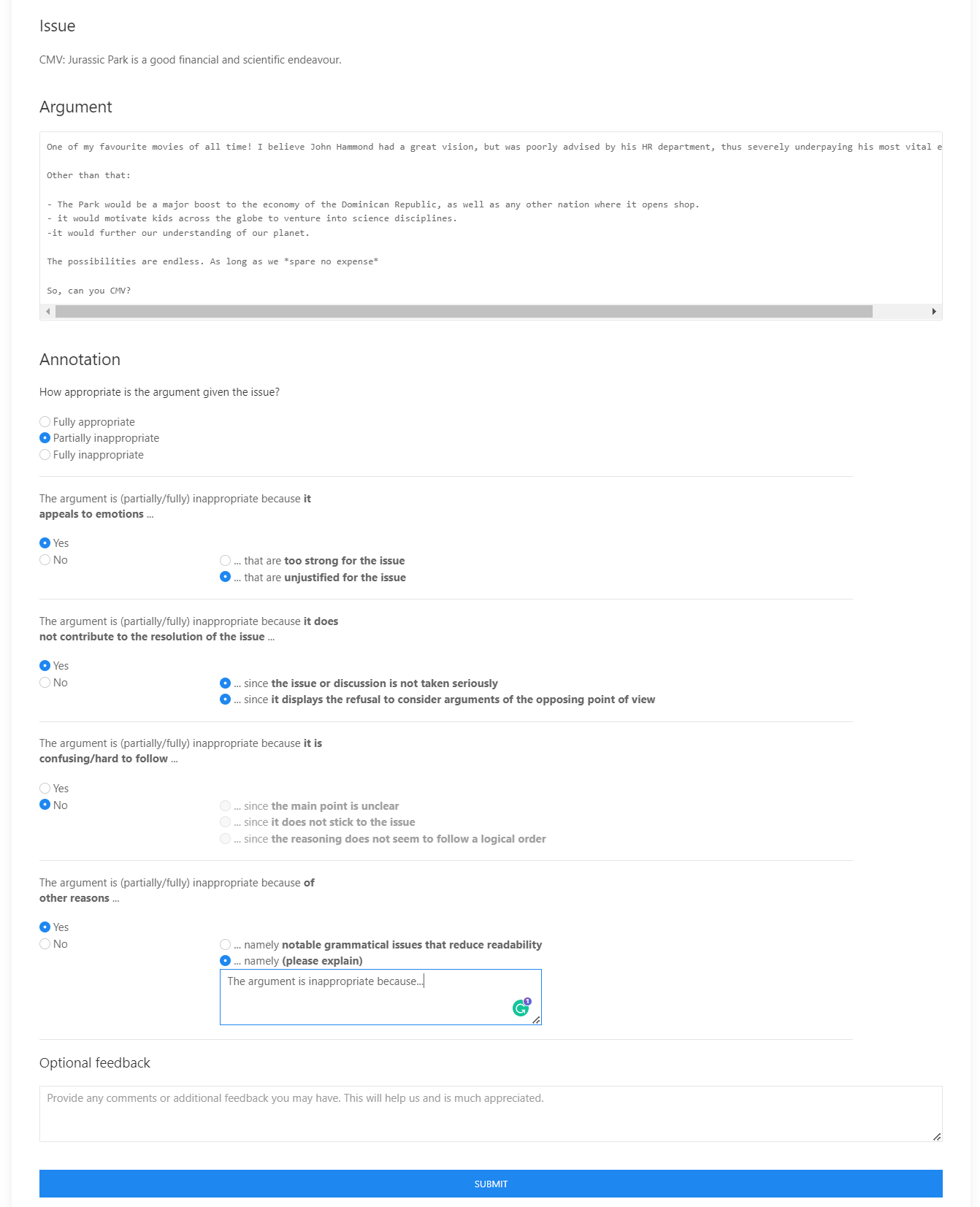}
\end{figure}

\clearpage

\section{GAQCorpus Correlations}
\label{sec:gaqcorpus}

\begin{table*}[h]
\small
\renewcommand{\arraystretch}{1}
\centering
\setlength{\tabcolsep}{1.2pt}
\begin{tabular*}{\linewidth}{l@{\quad}l@{\,\,}rr@{\quad}rrr@{\quad}rrr@{\quad}rrrr@{\quad}rrr}
\toprule
		&				&  \multicolumn{15}{c}{\bf Kendall's $\tau$ Correlation with Inappropriateness Dimensions} \\
						   \cmidrule(l@{-5pt}r@{1pt}){4-17}	
\bf Quality Dimension		&  \bf Description		&&  \bf In	& \bf TE	& \bf EI	& \bf ED	& \bf MC	& \bf MS	& \bf MO	& \bf MI	& \bf UM	& \bf MR	& \bf CR	& \bf OR	& \bf DO	& \bf RU\\
\midrule
Cogency    			& $P$ acceptable / relevant / sufficient			            && \bf .32      & .23       & .22	    & \bf .22   & .24	    & .14	    & .22	    & \bf .24	& \bf .18	& \bf .17	& \bf .11   & .09	    & .09	    & \bf .02	\\
Effectiveness     	& $a$ persuades target audience         						&& \bf .32      & .23	    & .22	    & .21	    & .25	    & \bf .15   & .22	    & .21	    & .16	    & .15	    & \bf .11   & \bf .10	& \bf .11	& .00	    \\
Reasonableness    	& $A$ acceptable / relevant / sufficient        				&& \bf .32      & \bf .25	& \bf .24	& \bf .22	& \bf .26	& \bf .15	& \bf .24	& .21	    & .16	    & .15	    & .10       & .09       & .10	    & .01	    \\
Overall quality    	& $a/A$ is of high quality									    && \bf .32      & .24	    & .23	    & \bf .22	& .24	    & \bf .15	& .23	    & .22	    & .17	    & .16	    & \bf .11	& \bf .10	& .10	    & .01	    \\
\bottomrule 
\end{tabular*}
\caption{Kendall's $\tau$ correlation of the argument \emph{quality dimensions} of \citet{ng:2020} with the mean ratings of the proposed \emph{appropriateness dimensions} (see Table~\ref{table-agreement-and-correlations} for the meaning of the acronyms). $P$ are all premises of an argument $a$ that is used within argumentation $A$. The highest value in each column is marked in bold.}
\label{table-correlations-gaq}
\end{table*}

\section{Corpus Statistics}
\label{sec:corpusstatistics}
\begin{table*}[h!]
\small
\renewcommand{\arraystretch}{0.9}
\centering
\setlength{\tabcolsep}{0.75pt}
\begin{tabular*}{\linewidth}{@{}	l@{\;\;}l@{\;\;}r@{\;\;}r r@{\quad} c@{\;\;}c	r@{\quad\;}	r@{\quad}rrr@{\quad}rrr@{\quad}rrrr@{\quad}rrr}
\toprule
		&				& \multicolumn{2}{c}{\bf (a) Count} && \multicolumn{2}{c}{\bf (b) Agree.}			&& \multicolumn{14}{c}{\bf (c) Kendall's $\tau$ Correlation} \\
						   \cmidrule(l@{-2pt}r@{-2pt}){3-4}	\cmidrule(l@{-2pt}r@{-2pt}){6-7}	      \cmidrule(l@{-2pt}r@{0pt}){9-22}	
    \multicolumn{2}{l}{\bf Dimension} 	& \bf Yes  & \bf No && \phantom{.} \bf Full & \phantom{.}\bf $\alpha$ && \bf In	& \bf TE	& \bf EI	& \bf ED	& \bf MC	& \bf MS	& \bf MO	& \bf MI	& \bf UM	& \bf MR	& \bf CR	& \bf OR	& \bf DO	& \bf RU\\
\midrule		
\bf In	& \bf Inappropriateness   	    & \bf 609   &  443	     && \phantom{..}52\tiny\%	  &.\bf 55    &&   	        & .48		& .32		& .38		& .59  	    & .40		& .45		& \bf .65	& .42		& .43  	    & .25  	    & .22		& .18		& .10    \\[1ex]%
\bf TE	& \bf Toxic Emotions    	    & 263       &  789	     && \phantom{..}80\tiny\%	  &.41        && .48	    & 	        & \bf .65	& \bf .79  	& .36  	    & .14		& .36		& .15		& .01		& .14		& .06		& .01		& .00  	    & .00    \\
EI		& Excessive Intensity   	    & 172       &  880	     && \phantom{..}84\tiny\%     &.35        && .32	    & .65	    &		    & .22		& .23		& .07   	& .24		& .12		& .01		& .10		& .07		& .02		& .02		& .01    \\
ED		& Emotional Deception   	    & 186       &  866	     && \phantom{..}84\tiny\%	  &.43        && .38	    & \bf.79    & .22	    &     	    & .31		& .15		& .27		& .11		& .03		& .10		& .04		& .01		& .01		& .00    \\[1ex]%
\bf MC	& \bf Missing Commitment        & 381       &  671	     && \phantom{..}69\tiny\%	  &.31        && .59	    & .36		& .23		& .31		&           & \bf .62	& \bf .78	& .22		& .08		& .21		& .04		& .01		& .01		& .00    \\
MS		& Missing Seriousness           & 135       &  917	     && \phantom{..}90\tiny\%	  &.\bf 55    && .40	    & .14		& .07		& .15		& .62		&           & .12   	& .15		& .10		& .17		& .01		& .01		& .02		& .01    \\
MO		& Missing Openness	            & 326       &  726	     && \phantom{..}70\tiny\%	  &.17        && .45	    & .36		& .24		& .27		& \bf .78   & .12		&           & .17		& .05		& .16		& .06		& .00		& .01		& .01    \\[1ex]%
\bf MI	& \bf Missing Intelligibility   & 460  		&  592	     && \phantom{..}62\tiny\%	  &.30        && \bf .65	& .15		& .12		& .11		& .22		& .15		& .17		& 	        & \bf .64	& \bf .65	& \bf .41	& .12   	& .14		& .01    \\
UM		& Unclear Meaning 		        & 300       &  752	     && \phantom{..}73\tiny\%	  &.18        && .42	    & .01   	& .01		& .03		& .08		& .10		& .05		& .64		&   	    & .17		& .21	    & .12       & .16		& .03    \\
MR		& Missing Relevance 	        & 297       &  755	     && \phantom{..}74\tiny\%	  &.25        && .43	    & .14		& .10  	    & .10		& .21		& .17		& .16		& \bf .65	& .17		&		    & .07		& .02		& .01	    & .01    \\
CR		& Confusing Reasoning	        & 135	    &  917	     && \phantom{..}87\tiny\%     &.17     	  && .25	    & .06		& .07		& .04		& .04		& .01		& .06		& .41		& .21		& .07		& 	        & .12		& .13		& .01    \\[1ex]%
\bf OR	& \bf Other Reasons   	        & 86        &  966	     && \phantom{..}92\tiny\%	  &.24        && .22	    & .01		& .02		& .01		& .01	    & .01		& .00		& .12		& .12		& .02		& .12		& 	        & \bf .87	& \bf.45 \\
DO		& Detrimental Orthography    	& 59        &  993	     && \phantom{..}95\tiny\%	  &.33        && .18	    & .00		& .02		& .01		& .01		& .02	    & .01   	& .14		& .16		& .01		& .13		& \bf .87	& 	        & .00    \\
RU		& Reason Unclassified	        & 28        &  \bf 1024  && \phantom{..}\bf 97\tiny\% &.01 	      && .10	    & .00		& .01		& .00		& .00		& .01		& .01	    & .01		& .03		& .01		& .01		& .45	    & .00       & 	   \\
\bottomrule
\end{tabular*}
\caption{Corpus statistics of the 1052 annotated \textbf{arguments} in the UKPConvArg2~\cite{habernal:2016b} corpus: (a) Counts of annotations for each inappropriateness dimension, when being aggregated conservatively (i.e., at least one annotator chose \emph{yes}). (b) \emph{Full} agreement and Krippendorff's $\alpha$ agreement of all three annotators. (c) Kendall's $\tau$ correlation between the 14 inappropriateness dimensions, averaged over the correlations of all annotators. The highest value in each column is marked in bold.}
\label{table-agreement-and-correlations-ukp}
\end{table*}

\begin{table*}[h!]
\small
\renewcommand{\arraystretch}{0.9}
\centering
\setlength{\tabcolsep}{0.75pt}
\begin{tabular*}{\linewidth}{@{}	l@{\;\;}l@{\;\;}r@{\;\;}r r@{\quad} c@{\;\;}c	r@{\quad\;}	r@{\quad}rrr@{\quad}rrr@{\quad}rrrr@{\quad}rrr}
\toprule
		&				& \multicolumn{2}{c}{\bf (a) Count} && \multicolumn{2}{c}{\bf (b) Agree.}			&& \multicolumn{14}{c}{\bf (c) Kendall's $\tau$ Correlation} \\
						   \cmidrule(l@{-2pt}r@{-2pt}){3-4}	\cmidrule(l@{-2pt}r@{-2pt}){6-7}	      \cmidrule(l@{-2pt}r@{0pt}){9-22}	
    \multicolumn{2}{l}{\bf Dimension} 	& \bf Yes  & \bf No && \phantom{.} \bf Full & \phantom{.}\bf $\alpha$ && \bf In	& \bf TE	& \bf EI	& \bf ED	& \bf MC	& \bf MS	& \bf MO	& \bf MI	& \bf UM	& \bf MR	& \bf CR	& \bf OR	& \bf DO	& \bf RU\\
\midrule		
\bf In	& \bf Inappropriateness   	    & \bf 271   &  267	     && \phantom{..}53\tiny\%	  &.22        &&   	        & .67		& .49		& .49		& .59  	    & .19		& .56		& .50		& .34  	    & .27  	    & .22		& .18		& .17       & .00\\[1ex]%
\bf TE	& \bf Toxic Emotions    	   	& 145       &  393	     && \phantom{..}76\tiny\%	  &\bf .28    && \bf .67	& 	        & \bf .69	& \bf .74  	& .33  	    & .04		& .32		& .04		& .00		& .06		& .01		& .01		& .01       & .00\\
EI		& Excessive Intensity   	    & 109       &  429	     && \phantom{..}80\tiny\%     &.15        && .49	    & .69	    &		    & .22		& .32		& .03   	& .31		& .03		& .00		& .05		& .02		& .03		& .03       & .00\\
ED		& Emotional Deception   		& 98        &  440	     && \phantom{..}83\tiny\%	  &\bf .28    && .49	    & \bf .74   & .22	    &     	    & .19		& .01		& .20		& .03		& .01		& .05		& .03		& .01		& .01       & .00\\[1ex]%
\bf MC	& \bf Missing Commitment       	& 174       &  364	     && \phantom{..}68\tiny\%	  &.07        && .58	    & .33		& .32		& .19		&           & \bf .35	& \bf .94	& .10		& .02		& .17		& .01		& .01		& .01	    & .00\\
MS		& Missing Seriousness           & 14        &  524	     && \phantom{..}97\tiny\%	  &.18        && .19	    & .04		& .03		& .01		& .35		&           & .11   	& .04		& .02		& .06		& .01		& .01		& .01	    & .00\\
MO		& Missing Openness	            & 166       &  372	     && \phantom{..}70\tiny\%	  &.07        && .56	    & .32		& .31		& .20		& \bf .94   & .11		&           & .09		& .01		& .16		& .02		& .01	    & .01	    & .00\\[1ex]%
\bf MI	& \bf Missing Intelligibility  	& 113  		&  425	     && \phantom{..}79\tiny\%	  &.10        && .50	    & .04		& .03		& .03		& .10		& .04		& .09		&   	    & \bf .65	& \bf .57	& \bf .41  	& .20		& .22       & .00\\
UM		& Unclear Meaning 		        & 66        &  472	     && \phantom{..}88\tiny\%	  &.03        && .34	    & .00	    & .00	    & .01	    & .02	    & .02	    & .01		& \bf .65   & 	    	& .06		& .17	    & .32       & .36       & .00\\
MR		& Missing Relevance 	        & 51        &  487	     && \phantom{..}91\tiny\%	  &.09        && .27	    & .06		& .05		& .05		& .17		& .06		& .16		& .57	    & .06	    &   	    & .01	    & .01       & .01       & .00\\
CR		& Confusing Reasoning	        & 18	    &  520	     && \phantom{..}97\tiny\%     &.09     	  && .22	    & .01		& .02		& .03		& .01		& .01	    & .02		& .41	    & .17	    & .01		&           & .17       & .19       & .00\\[1ex]%
\bf OR	& \bf Other Reasons   	       	& 11        &  527	     && \phantom{..}98\tiny\%	  &.23        && .18	    & .01		& .03		& .01		& .01	    & .01	    & .01	    & .20	    & .32		& .01	    & .17		&   	    & \bf .96   & .00\\
DO		& Detrimental Orthography      	& 10        &  528	     && \phantom{..}98\tiny\%	  &.24        && .17	    & .01		& .03		& .01		& .01	    & .01	    & .01	    & .22	    & .36		& .01	    & .19		& \bf .96   &   	    & .00\\
RU		& Reason Unclassified	        & 1         &  \bf 537   && \phantom{..}\bf 100\tiny\%&.00 	      && .00	    & .00		& .00		& .00		& .00		& .00		& .00	    & .00		& .00		& .00		& .00		& .00		& .00	    &    \\
\bottomrule
\end{tabular*}
\caption{Corpus statistics of the 538 annotated \textbf{arguments} in the GAQCorpus~\cite{ng:2020}: (a) Counts of annotations for each inappropriateness dimension, when being aggregated conservatively (i.e., at least one annotator chose \emph{yes}). (b) \emph{Full} agreement and Krippendorff's $\alpha$ agreement of all three annotators. (c) Kendall's $\tau$ correlation between the 14 inappropriateness dimensions, averaged over the correlations of all annotators. The highest value in each column is marked in bold.}
\label{table-agreement-and-correlations-debates}
\end{table*}

\begin{table*}[h!]
\small
\renewcommand{\arraystretch}{0.9}
\centering
\setlength{\tabcolsep}{0.75pt}
\begin{tabular*}{\linewidth}{@{}	l@{\;\;}l@{\;\;}r@{\;\;}r r@{\quad} c@{\;\;}c	r@{\quad\;}	r@{\quad}rrr@{\quad}rrr@{\quad}rrrr@{\quad}rrr}
\toprule
		&				& \multicolumn{2}{c}{\bf (a) Count} && \multicolumn{2}{c}{\bf (b) Agree.}			&& \multicolumn{14}{c}{\bf (c) Kendall's $\tau$ Correlation} \\
						   \cmidrule(l@{-2pt}r@{-2pt}){3-4}	\cmidrule(l@{-2pt}r@{-2pt}){6-7}	      \cmidrule(l@{-2pt}r@{0pt}){9-22}	
    \multicolumn{2}{l}{\bf Dimension} 	& \bf Yes  & \bf No && \phantom{.} \bf Full & \phantom{.}\bf $\alpha$ && \bf In	& \bf TE	& \bf EI	& \bf ED	& \bf MC	& \bf MS	& \bf MO	& \bf MI	& \bf UM	& \bf MR	& \bf CR	& \bf OR	& \bf DO	& \bf RU\\
\midrule		
\bf In	& \bf Inappropriateness   	    & \bf 279  &  221	    && \phantom{..}47\tiny\%	    &.33        &&   	    & .68		& .44		& .54		& .55  	    & .30		& .44		& .59		& .36		& .42  	    & .20  	    & .15		& .00		& .00    \\[1ex]%
\bf TE	& \bf Toxic Emotions    	    & 171       &  329	    && \phantom{..}71\tiny\%	    &.34        && \bf .68	& 	        & \bf .63	& \bf .78  	& .37  	    & .11		& .35		& .18		& .04		& .12		& .11		& .00		& .00		& .00    \\
EI		& Excessive Intensity   	    & 111       &  389	    && \phantom{..}78\tiny\%        &.25        && .44	    & .63	    &		    & .18		& .26		& .07   	& .25		& .16		& .04		& .09		& .16		& .03		& .00		& .00    \\
ED		& Emotional Deception   		& 133       &  367	    && \phantom{..}75\tiny\%	    &.30        && .54	    & \bf.78    & .18	    &     	    & .28		& .09		& .25		& .10		& .01		& .08		& .04		& .01		& .00		& .00    \\[1ex]%
\bf MC	& \bf Missing Commitment        & 177       &  323	    && \phantom{..}66\tiny\%	    &.09        && .55	    & .37		& .26		& .28		&           & \bf .54	& \bf .79	& .25		& .04		& .20		& .12		& .06		& .00		& .00    \\
MS		& Missing Seriousness           & 32        &  468	    && \phantom{..}94\tiny\%	    &\bf .40    && .30	    & .11		& .07		& .09		& .54		&          	& .07       & .11		& .01		& .13		& .05		& .03		& .00		& .00    \\
MO		& Missing Openness	            & 165       &  335	    && \phantom{..}67\tiny\%	    &.00        && .44	    & .35		& .25		& .25		& \bf .79   & .07		&           & .25		& .06		& .19		& .11		& .03		& .00		& .00    \\[1ex]%
\bf MI	& \bf Missing Intelligibility   & 189  		&  311	    && \phantom{..}64\tiny\%	    &.11        && .59	    & .18		& .16		& .10		& .25		& .11		& .25		& 	        & \bf .60	& \bf .72	& \bf .33	& .11   	& .00		& .00    \\
UM		& Unclear Meaning 		        & 90        &  410	    && \phantom{..}82\tiny\%	    &.07        && .36	    & .04   	& .04		& .01		& .04		& .01		& .06		& .60		&   	    & .14		& .16	    & \bf .17   & .00		& .00    \\
MR		& Missing Relevance 	        & 150       &  350	    && \phantom{..}71\tiny\%	    &.04        && .42	    & .12		& .09  	    & .08		& .20		& .13		& .19		& \bf .72	& .14		&		    & .04		& .00		& .00	    & .00    \\
CR		& Confusing Reasoning	        & 20	    &  480	    && \phantom{..}96\tiny\%        &.04     	&& .20	    & .11		& .16		& .04		& .12		& .05		& .11		& .33		& .16		& .04		& 	        & .14		& .00		& .00    \\[1ex]%
\bf OR	& \bf Other Reasons   	        & 11        &  489	    && \phantom{..}98\tiny\%	    &.08        && .15	    & .00		& .03		& .01		& .06	    & .03		& .03		& .11		& .17		& .00		& .14		& 	        & .00	    & .00 \\
DO		& Detrimental Orthography    	& 8         &  492	    && \phantom{..}98\tiny\%	    &.11        && .00	    & .00		& .00		& .00		& .00		& .00		& .00   	& .00		& .00		& .00		& .00		& .00	    & 	        & .00    \\
RU		& Reason Unclassified	        & 3         &  \bf 497  && \phantom{..}\bf 99\tiny\%    &.00 	    && .00	    & .00		& .00		& .00		& .00		& .00		& .00	    & .00		& .00		& .00		& .00		& .00		& .00		&        \\
\bottomrule
\end{tabular*}
\caption{Corpus statistics of the 500 annotated \textbf{forum posts} in the GAQCorpus~\cite{ng:2020}: (a) Counts of annotations for each inappropriateness dimension, when being aggregated conservatively (i.e., at least one annotator chose \emph{yes}). (b) \emph{Full} agreement and Krippendorff's $\alpha$ agreement of all three annotators. (c) Kendall's $\tau$ correlation between the 14 inappropriateness dimensions, averaged over the correlations of all annotators. The highest value in each column is marked in bold.}
\label{table-agreement-and-correlations-qa}
\end{table*}

\begin{table*}[t!]
\small
\renewcommand{\arraystretch}{0.9}
\centering
\setlength{\tabcolsep}{0.75pt}
\begin{tabular*}{\linewidth}{@{}	l@{\;\;}l@{\;\;}r@{\;\;}r r@{\quad} c@{\;\;}c	r@{\quad\;}	r@{\quad}rrr@{\quad}rrr@{\quad}rrrr@{\quad}rrr}
\toprule
		&				& \multicolumn{2}{c}{\bf (a) Count} && \multicolumn{2}{c}{\bf (b) Agree.}			&& \multicolumn{14}{c}{\bf (c) Kendall's $\tau$ Correlation} \\
						   \cmidrule(l@{-2pt}r@{-2pt}){3-4}	\cmidrule(l@{-2pt}r@{-2pt}){6-7}	      \cmidrule(l@{-2pt}r@{0pt}){9-22}	
    \multicolumn{2}{l}{\bf Dimension} 	& \bf Yes  & \bf No && \phantom{.} \bf Full & \phantom{.}\bf $\alpha$ && \bf In	& \bf TE	& \bf EI	& \bf ED	& \bf MC	& \bf MS	& \bf MO	& \bf MI	& \bf UM	& \bf MR	& \bf CR	& \bf OR	& \bf DO	& \bf RU\\
\midrule		
\bf In	& \bf Inappropriateness   	    & \bf 23    &  78	     && \phantom{..}79\tiny\%	    & .44        &&   	    & .78		& .56		& .63		& .00  	    & .00		& .00		& .52		& .00		& .48  	    & .00   & -		& -		& -    \\[1ex]%
\bf TE	& \bf Toxic Emotions    	    & 15        &  86	     && \phantom{..}87\tiny\%	    & .41        && \bf .78	& 	        & \bf .73  	& \bf .83  	& .00		& .00		& .00		& .16		& .00		& .17		& .00	& -  	& -     & -    \\
EI		& Excessive Intensity   	    & 10        &  91	     && \phantom{..}90\tiny\%       & .31        && .56	    & .73	    &		    & .43		& .00		& .00   	& .00		& .14		& .00		& .15		& .00	& -		& -		& -    \\
ED		& Emotional Deception   		& 10        &  91	     && \phantom{..}92\tiny\%	    & \bf .50    && .63	    & \bf .83   & .43	    &     	    & .00		& .00		& .00		& .05		& .00		& .06		& .00	& -		& -		& -    \\[1ex]%
\bf MC	& \bf Missing Commitment        & 3         &  98	     && \phantom{..}97\tiny\%	    & .01        && .00	    & .00		& .00		& .00		&           & .00   	& .00  	    & .00		& .00		& .00		& .00	& -		& -		& -    \\
MS		& Missing Seriousness           & 2         &  99	     && \phantom{..}98\tiny\%	    & .00        && .00	    & .00		& .00		& .00		& .00		&           & .00      	& .00		& .00		& .00		& .00	& -		& -		& -    \\
MO		& Missing Openness	            & 1         &  \bf 100	 && \phantom{..}\bf 99\tiny\%	& .00        && .00	    & .00		& .00		& .00		& .00       & .00		&           & .00		& .00       & .00		& .00	& -		& -		& -    \\[1ex]%
\bf MI	& \bf Missing Intelligibility   & 12  		&  89	     && \phantom{..}88\tiny\%	    & .16        && .52	    & .16		& .14		& .05		& .00		& .00		& .00		& 	        & .00   	& \bf .92	& .00	& -   	& -		& -    \\
UM		& Unclear Meaning 		        & 3         &  98	     && \phantom{..}97\tiny\%	    & .01        && .00	    & .00   	& .00		& .00		& .00		& .00		& .00		& .00		&           & .00		& .00   & -     & -		& -    \\
MR		& Missing Relevance 	        & 10        &  91	     && \phantom{..}90\tiny\%	    & .20        && .48	    & .17		& .15  	    & .06		& .00		& .00		& .00		& \bf .92	& .00		&		    & .00  	& -		& -	    & -    \\
CR		& Confusing Reasoning	        & 1	        &  \bf 100	 && \phantom{..}\bf 99\tiny\%   & .00    	 && .00	    & .00		& .00		& .00		& .00		& .00		& .00		& .00		& .00		& .00		&    	& -		& -		& -    \\[1ex]%
\bf OR	& \bf Other Reasons   	        & 0         &  0	     && -	                        & -	         && -		& -     & -		& -	    & -		& -		& -		& -		& -		& -		& -     & -     & -   & -    \\
DO		& Detrimental Orthography    	& 0         &  0	     && -	                        & - 	     && -		& -		& -		& -		& -	    & -   	& -		& -		& -		& -		& - 	& -     & -   & -    \\
RU		& Reason Unclassified	        & 0         &  0         && -	                        & - 	     && -		& -		& -		& -		& -		& -	    & -		& -		& -		& -		& -		& -	    & -   & -    \\
\bottomrule
\end{tabular*}
\caption{Corpus statistics of the 101 annotated \textbf{reviews} in the GAQCorpus~\cite{ng:2020}: (a) Counts of annotations for each inappropriateness dimension, when being aggregated conservatively (i.e., at least one annotator chose \emph{yes}). (b) \emph{Full} agreement and Krippendorff's $\alpha$ agreement of all three annotators. (c) Kendall's $\tau$ correlation between the 14 inappropriateness dimensions, averaged over the correlations of all annotators. The highest value in each column is marked in bold.}
\label{table-agreement-and-correlations-reviews}
\end{table*}

\clearpage

\end{document}